\documentclass[conference]{IEEEtran}
\IEEEoverridecommandlockouts
\usepackage{amsmath,amssymb,amsfonts}
\usepackage{algorithmic}
\usepackage{graphicx}
\usepackage{tabularx}
\usepackage{array}
\usepackage{multirow}
\usepackage{subfig}
\usepackage{textcomp}
\usepackage{placeins}
\usepackage{xcolor}
\usepackage{tabularx}
\usepackage{float} 
\usepackage{soul}
\usepackage{colortbl} % Add this in the preamble
\usepackage{xcolor}   % Allows custom colors

\usepackage{fancyhdr}

\def\BibTeX{{\rm B\kern-.05em{\sc i\kern-.025em b}\kern-.08em
    T\kern-.1667em\lower.7ex\hbox{E}\kern-.125emX}}

\makeatother

\begin{document}

\title{A Model-Mediated Stacked Ensemble Approach for Depression Prediction Among Professionals}

\author{
\IEEEauthorblockN{Md. Mortuza Ahmmed}
\IEEEauthorblockA{\textit{Associate Professor} \\
\textit{Department of Mathematics}\\
\textit{American International}\\
\textit{University-Bangladesh}\\
Dhaka, Bangladesh \\
mortuza@aiub.edu}
\and
\IEEEauthorblockN{Abdullah Al Noman}
\IEEEauthorblockA{\textit{Undergraduate Student} \\
\textit{Department of Computer Science}\\
\textit{American International}\\
\textit{University-Bangladesh}\\
Dhaka, Bangladesh \\
22-46609-1@student.aiub.edu}
\and
\IEEEauthorblockN{Mahin Montasir Afif}
\IEEEauthorblockA{\textit{Undergraduate Student} \\
\textit{Department of Computer Science}\\
\textit{American International}\\
\textit{University-Bangladesh}\\
Dhaka, Bangladesh \\
22-46573-1@student.aiub.edu}
\and
\IEEEauthorblockN{K. M. Tahsin Kabir}
\IEEEauthorblockA{\textit{Lecturer} \\
\textit{Department of Computer Science}\\
\textit{and Engineering}\\
\textit{Asian Univerisy of Bangladesh}\\
Dhaka, Bangladesh \\
tahsin@aub.ac.bd}
\and
\IEEEauthorblockN{Md. Mostafizur Rahman}
\IEEEauthorblockA{\textit{Associate Professor} \\
\textit{Department of Mathematics}\\
\textit{American International} \\
\textit{University-Bangladesh}\\
Dhaka, Bangladesh\\
mostafiz.math@aiub.edu}

\and
\IEEEauthorblockN{Mufti Mahmud}
\IEEEauthorblockA{\textit{Professor, Information and Computer} \\
\textit{Science Department}\\
\textit{King Fahd University of Petroleum} \\
\textit{and Minerals (KFUPM)}\\
Saudi Arabia\\
mufti.mahmud@kfupm.edu.sa}

}

\maketitle

\begin{abstract}
Depression is a significant mental health concern, particularly in professional environments where work-related stress, financial pressure, and lifestyle imbalances contribute to deteriorating well-being. Despite increasing awareness, researchers and practitioners face critical challenges in developing accurate and generalizable predictive models for mental health disorders. Traditional classification approaches often struggle with the complexity of depression, as it is influenced by multifaceted, interdependent factors, including occupational stress, sleep patterns, and job satisfaction. This study addresses these challenges by proposing a stacking-based ensemble learning approach to improve the predictive accuracy of depression classification among professionals. Depression Professional Dataset has been collected from Kaggle. The dataset comprises demographic, occupational, and lifestyle attributes that influence mental well-being. Our stacking model integrates multiple base learners with a logistic regression mediated model, effectively capturing diverse learning patterns. The experimental results demonstrate that the proposed model achieves high predictive performance, with an accuracy of 99.64\% on training data and 98.75\% on testing data with a precision, recall, and F1-score all exceeding 98\%. These findings highlight the effectiveness of ensemble learning in mental health analytics and underscore its potential for early detection and intervention strategies. 

\end{abstract}


\maketitle

\section{\label{sec: intro}Introduction}
In recent years, the prevalence of psychological disorders such as depression, anxiety, and stress has increased significantly due to the fast-paced nature of modern life. As a result, this has driven considerable interest in applying machine learning (ML) techniques for early detection, accurate diagnosis, and effective treatment prediction of mental health issues. These technologies offer the potential to analyze large volumes of behavioral and physiological data, uncover hidden patterns, and provide insights that might be missed by traditional diagnostic methods. With the growing accessibility of digital health records and wearable devices, ML-based tools are becoming increasingly viable for real-world clinical applications.
Several studies have focused on predicting psychological distress using questionnaire-based data and various ML algorithms. One such work \cite{PRIYA20201258} employed the Depression, Anxiety and Stress Scale (DASS-21) to gather data from individuals of diverse backgrounds and applied five ML models to classify the severity of psychological conditions into five levels. Among the algorithms tested, Random Forest emerged as the most effective, particularly in handling imbalanced class distributions through F1-score and specificity evaluations. A broader view is offered in review studies such as that by \cite{aleem2022depression}, which systematically categorized ML techniques into classification, deep learning, and ensemble models for depression diagnosis. These models typically follow a pipeline involving data preprocessing, feature selection, classifier training, and performance evaluation. The study emphasized the growing potential of ML to outperform traditional diagnostic methods and presented insights into both the strengths and limitations of existing approaches.

Further in-depth analysis was presented by \cite{ZULFIKER2021100044} , who tested six machine learning classifiers using socio-demographic and psychosocial variables to detect depression. With SMOTE used to address class imbalance and feature selection techniques such as SelectKBest, mRMR, and Boruta applied, the AdaBoost classifier combined with SelectKBest yielded the best accuracy of 92.56\%, demonstrating the effectiveness of tailored feature selection in enhancing predictive accuracy. While survey-based approaches provide valuable insights, other research has explored the use of neuroimaging data for depression analysis. Studies such as that by \cite{PATEL2016115} utilized functional and structural imaging data to distinguish between depressed and non-depressed individuals and predict treatment outcomes.

Parallel to this, study into treatment outcome prediction using ML has gained momentum. A notable study by \cite{chekroud2016cross} trained a model on the STAR*D dataset and externally validated it on the COMED trial, showing statistically significant predictions for remission in patients treated with citalopram and escitalopram, though with moderate accuracies around 60–65\%. These results suggest that ML can assist in personalized treatment planning, although model generalizability remains a challenge. Expanding on this, a meta-analysis and systematic review by \cite{LEE2018519} synthesized findings across multiple studies, reporting an overall predictive accuracy of 82\% for therapeutic outcomes in mood disorders using ML. Models that integrated multi-modal data (e.g., neuroimaging, genomics, and clinical features) achieved significantly better accuracy than those relying on single data types. However, issues related to study heterogeneity, retrospective designs, and lack of standardization across ML pipelines were noted as major limitations.

Another meta-analysis by \cite{Sajjadian2021} focused specifically on major depressive disorder (MDD) and found that high-quality studies had a lower mean accuracy (~63\%) compared to others (~75\%), suggesting a potential overestimation of ML performance in lower-rigor settings. Moreover, the ability to predict treatment resistance surpassed that of predicting remission or response, indicating varying effectiveness depending on the clinical target.
A more technical perspective was explored in a study by \cite{LI2019101696} that evaluated EEG-based depression recognition using feature extraction methods (e.g., power spectral density, Hjorth parameters) and compared ensemble learning with deep learning models. This study demonstrated the value of objective biosignal-based methods in reducing diagnostic subjectivity and enhancing classification performance through sophisticated signal processing and model tuning. Similarly, \cite{Lee2022} explored early depression diagnosis using EEG data and ML, reinforcing the importance of physiological signals as biomarkers for mental health conditions.

In a recent and notably robust study, \cite{Richter2020} presented a machine learning-based behavioral analysis approach to differentiate between anxiety and depression. Using a comprehensive cognitive-emotional test battery and custom-built ML models, the study achieved over 70\% accuracy in identifying distinct symptom patterns, laying the groundwork for improved diagnostic instruments and more personalized treatment strategies. Among multi-class classification efforts, \cite{KUMAR20201989} demonstrated one of the most accurate models for assessing depression, anxiety, and stress levels, achieving high performance across all three categories, and offering a reliable multiclass prediction model. Finally, \cite{Jain_2022} combined natural language processing (NLP) with ML to analyze depression and suicide risk through social media data, offering an innovative approach to mental health surveillance through digital footprints.

Table 1 shows the overview of the literature review. Compared to these studies, our research uniquely focuses on predicting psychological disorders using demographic, occupational, and lifestyle attributes that influence mental well-being while applying multiple ML algorithms without relying on external clinical or biosignal data. While prior works emphasized accuracy or feature importance, our study also emphasizes comparative performance evaluation using key metrics like accuracy, precision, recall, and F1-score, offering a transparent, balanced view of model reliability. Moreover, unlike neuroimaging studies that require high computational resources and clinical expertise, our approach remains accessible and scalable for educational institutions or public health surveys. Thus, our study contributes to the field by offering a replicable, lightweight model for early psychological disorder detection, particularly relevant in low-resource settings.

\begin{table}[htbp]
\caption{Summary of ML Studies on Mental Health Diagnosis}
\label{tab:ml_summary}
\centering
\scriptsize % smaller font to fit narrow IEEE column
\setlength{\tabcolsep}{3pt} % reduce space between columns
\begin{tabular}{|>{\raggedright\arraybackslash}p{0.5cm}|
                >{\raggedright\arraybackslash}p{1.4cm}|
                >{\raggedright\arraybackslash}p{1.5cm}|
                >{\raggedright\arraybackslash}p{3.3cm}|}
\hline
\textbf{Ref.} & \textbf{Accuracy} & \textbf{Dataset} & \textbf{Research Gap} \\
\hline
\cite{PRIYA20201258} & ~88\% (RF) & DASS-21 & Limited generalizability; self-report only \\
\hline
\cite{ZULFIKER2021100044} & 92.6\% (AdaBoost) & Survey data & Self-reporting; overfitting risk; limited validation \\
\hline
\cite{chekroud2016cross} & 60--65\% & STAR*D, COMED & Moderate accuracy; poor generalization \\
\hline
\cite{LEE2018519} & 82\% & Multiple datasets & Heterogeneous data; retrospective limits \\
\hline
\cite{Sajjadian2021} & 63--75\% & Various datasets & Bias in low-rigor studies; varies by target \\
\hline
\cite{LI2019101696} & ~85\% (Ensemble) & EEG biosignals & Diagnostic subjectivity; small sample \\
\hline
\cite{Lee2022} & 95.6\% (LDA) & Clinical records & Limited diversity; low interpretability \\
\hline
\cite{PATEL2016115} & 94.7\% (SVM) & DASS-21 & Class imbalance; questionnaire-based only \\
\hline
\cite{Richter2020} & ~92\% (CNN+RF) & Behavioral video & Complex acquisition; real-time challenges \\
\hline
\cite{Jain_2022} & ~89\% (NB) & Social media & Labeling bias; privacy and ethics \\
\hline
\end{tabular}
\end{table}

\section{\label{sec:Method}Methodology}
In this section, we present the detailed pipeline of our proposed ensemble model, which was designed to enhance depression prediction among professionals. The methodological framework comprises data collection, data preprocessing, feature selection and development of a stacking ensemble model. The methodology overview is shown in figure 1.

\begin{figure}
    \centering
    \includegraphics[width=1\linewidth]{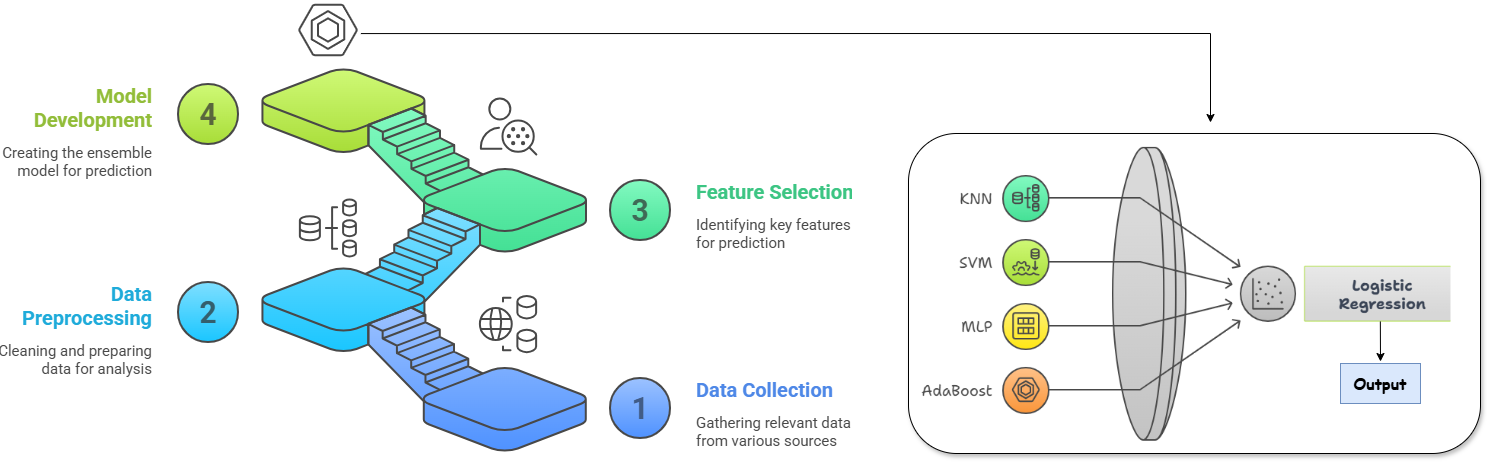}
    \caption{Methodology overview of Stacking Ensemble Model}
    \label{fig:enter-label}
\end{figure}

\FloatBarrier

\subsection{Dataset Overview}
The Depression Professional Dataset\cite{sharma2023depression} has been collected from Kaggle as part of a comprehensive survey aimed at understanding the factors contributing to depression risk among adults. It was gathered during an anonymous survey conducted between January and June 2023 across various cities, targeting individuals from diverse backgrounds and professions aged between 18 to 60. The dataset inspects the relationship between mental health and various demographic, lifestyle, and work-related factors. It includes information on gender, age, work pressure, job satisfaction, sleep duration, dietary habits, financial stress, work hours, and mental health indicators such as depression, suicidal thoughts, and family history of mental illness. It illustrates how lifestyle and work conditions influence mental health and the impact of work-life balance.

\subsection{Data Preprocessing}
\begin{figure}[htbp]
    \centering
    \includegraphics[width=1\linewidth]{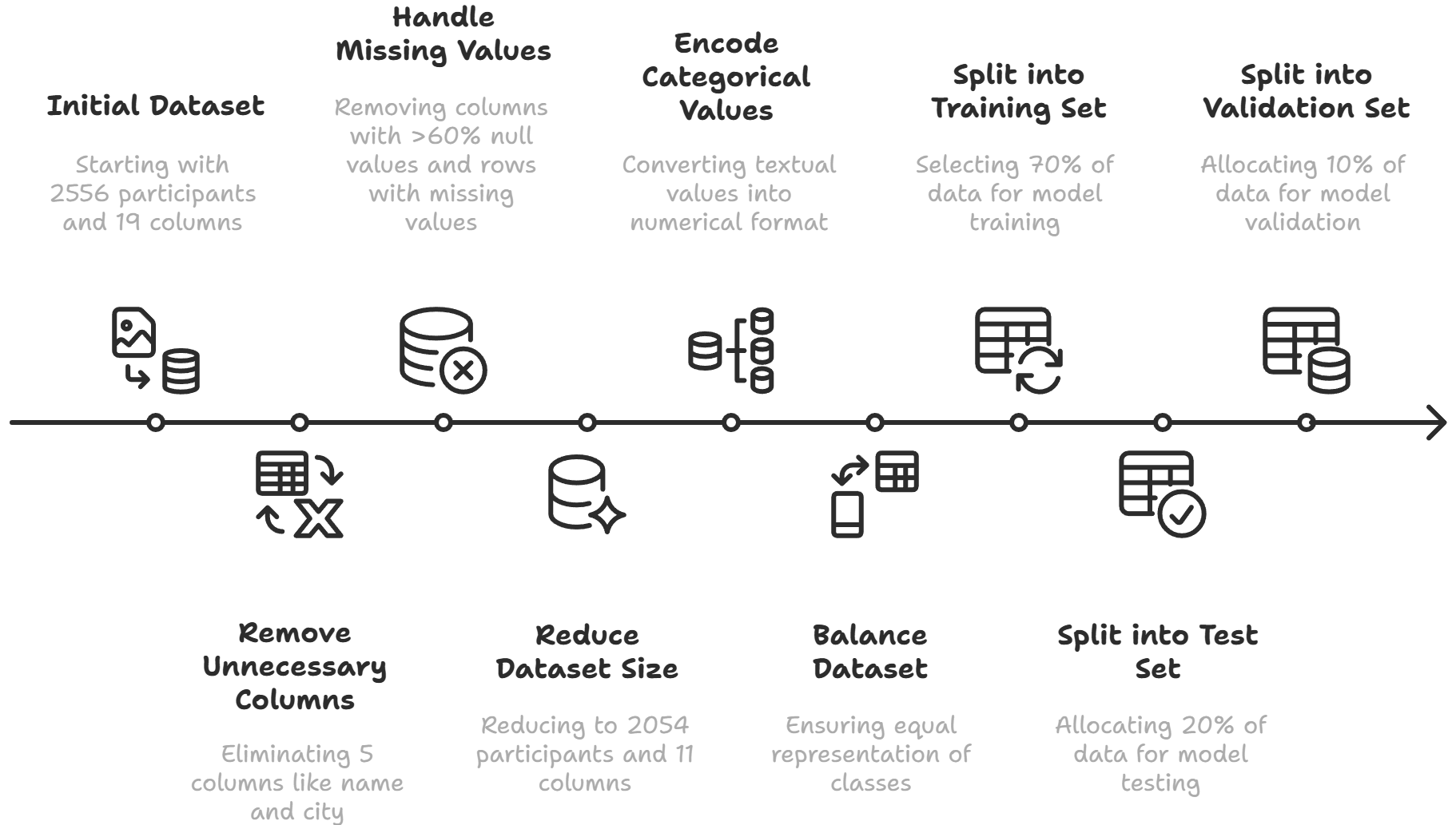}
    \caption{Overview of data preprocessing techniques}
    \label{fig:Preprocessing workflow}
\end{figure}

Figure \ref{fig:Preprocessing workflow} shows the full preprocessing steps  in detail. Before the model could be fitted, the dataset needed to be preprocessed. The raw dataset contains data on 2556 participants in total, with 19 columns. Initially, five unnecessary columns, such as participant’s name, type, and city were removed. To handle missing values, three columns with more than 60\% null values and rows with missing values were eliminated. After cleaning, the dataset was reduced to 2054 participants with 11 columns including the target feature. Encoding was used for categorical textual values. The dataset was then balanced for binary classification.
From the preprocessed dataset, 70\% of the samples were randomly selected as the training set for building the model and performing feature selection. The remaining 30\% was split into 20\% for the test set and 10\% for the validation set.

\subsection{Feature Selection}
Feature selection \cite{SosaCabrera2024} is a key preprocessing step that helps identify the most relevant features in the dataset. In this work, the chi-square ($\chi^2$) test was used to measure the statistical relationship between each feature and the target variable. Features showing strong significance were kept for modeling to enhance performance and reduce noise.

\subsubsection{Chi-Square Test:}
\begin{equation}
    X^2 = \sum \frac{(A - B)^2}{B}
\end{equation}

The Chi-Square test \cite{Byeon2017CHAID} compares observed counts \( A \) with expected counts \( B \) in categorical data. A higher test statistic indicates a stronger relationship between a feature and the target.

\begin{equation}
    \nu = (m - 1)(n - 1)
\end{equation}

\paragraph{Degrees of Freedom} 
The degrees of freedom \( \nu \) \cite{PONOMARENKO2013290} depend on the size of the contingency table and are calculated as one less than the number of rows times one less than the number of columns.

\paragraph{Cumulative Distribution Function (CDF) \cite{zhang2024functionallinearregressioncumulative}}  
The p-value is the probability of observing a test statistic as extreme as \( X^2_{\text{calc}} \), assuming the null hypothesis is true. It is computed as:

\begin{equation}
    p = P(X^2 \geq X^2_{\text{calc}}) = 1 - G(X^2_{\text{calc}}; \nu)
\end{equation}

\subsection{Development of Stacking Ensemble Model}
A stacking ensemble model was developed using K-Nearest Neighbors (KNN), Support Vector Machine (SVM), Multi-Layer Perceptron (MLP), and AdaBoost as base models, with Logistic Regression as the mediated model. The main objective was to evaluate the predictive performance of machine learning algorithms.

K-Nearest Neighbors (KNN) \cite{Guo2003} is a memory-based model that classifies a new instance based on the majority class of its nearest neighbors, using a distance metric. 

\begin{equation}
    d(x, x') = \sqrt{\sum_{i=1}^{n}(x_i - x'_i)^2}
\end{equation}

Here, $\mathbf{x}$ is a feature vector which representing the first data point, and let $\mathbf{x'}$ is another feature vector represents the second data point, which is to be compared with $\mathbf{x}$. The value of the $i$-th feature in vector $\mathbf{x}$ is denoted as $x_i$, while the corresponding value in vector $\mathbf{x'}$ is denoted as $x_i'$.  $n$ represent the total number of features in each data point. The Euclidean distance between the two points $\mathbf{x}$ and $\mathbf{x'}$ is denoted as $d(\mathbf{x}, \mathbf{x'})$.

Support Vector Machine (SVM) \cite{Suthaharan2016} is a supervised learning algorithm designed to identify the optimal hyperplane that separates instances of different classes with the maximum possible margin. The associated optimization problem can be expressed as:

\begin{equation}
\min_{\theta, \beta, \epsilon} \left( \frac{1}{2} |\theta|^2 + \lambda \sum_{j=1}^{m} \epsilon_j \right)
\end{equation}

In this formulation, $\theta$ denotes the weight vector that determines the orientation of the separating hyperplane, and $\beta$ represents the bias term that shifts the hyperplane from the origin. The variable $\epsilon_j$ is a slack variable for the $j$-th training instance, allowing some flexibility for misclassification or margin violation. The regularization parameter $\lambda$ balances the trade-off between maximizing the margin and reducing the classification errors by penalizing the slack variables. Here, $m$ refers to the total number of training samples. Minimizing the squared norm $|\theta|^2$ corresponds to maximizing the margin between the classes.

A Multi-Layer Perceptron (MLP) \cite{MURTAGH1991183} is a form of feedforward neural network that consists of one or more hidden layers. Each neuron processes its inputs by computing a weighted sum followed by the application of an activation function.

\begin{equation}
u = \sum \theta_j x_j + \gamma
\end{equation}

\begin{equation}
\phi(u) = \frac{1}{1 + e^{-u}}
\end{equation}

Here, $\theta_j$ refers to the weight associated with the $j$-th input feature $x_j$, and $\gamma$ is the bias term. The expression $u$ denotes the linear combination of inputs, which is then passed through an activation function $\phi(u)$—in this case, the sigmoid function.

\begin{equation}
\mathcal{J} = -\frac{1}{m} \sum_{k=1}^{m} [t_k \log(\hat{t}_k) + (1 - t_k) \log(1 - \hat{t}_k)]
\end{equation}

The variable $\mathcal{J}$ represents the cost function, specifically the binary cross-entropy loss, where $m$ is the total number of samples. The true label for each example is denoted by $t_k$, while $\hat{t}_k$ represents the predicted probability output.

\begin{equation}
\Theta = \Theta - \alpha \cdot \frac{\partial \mathcal{J}}{\partial \Theta}
\end{equation}

In the above, $\Theta$ indicates the weight vector prior to the update, and $\alpha$ is the learning rate. The gradient of the loss $\mathcal{J}$ with respect to the weights is given by $\frac{\partial \mathcal{J}}{\partial \Theta}$, which guides the weight adjustment during training.

AdaBoost \cite{Schapire2013} is an ensemble learning algorithm that constructs a powerful classifier by combining several weak learners. It iteratively updates the weights of training instances, placing greater emphasis on those that were previously misclassified.

\begin{equation}
    F(x) = \text{sign} \left( \sum_{k=1}^{K} \beta_k \cdot g_k(x) \right)
\end{equation}

In this expression, \( F(x) \) represents the final aggregated (strong) classifier. The ensemble consists of \( K \) weak classifiers. Each weak learner \( g_k(x) \) contributes to the final prediction with an associated weight \( \beta_k \). The \( \text{sign} \) function determines the final output by returning either \( +1 \) or \( -1 \) depending on the sign of the sum.

\begin{equation}
    \beta_k = \frac{1}{2} \ln \left( \frac{1 - \epsilon_k}{\epsilon_k} \right)
\end{equation}

Here, \( \beta_k \) indicates the importance (or influence) of the \( k \)-th weak classifier in the final decision, while \( \epsilon_k \) denotes the weighted classification error made by that particular weak learner. A lower error results in a higher weight, giving more reliable classifiers a stronger influence on the final result.

Logistic regression \cite{sperandei2014understanding} can function as a meta-classifier in a stacking ensemble, where it consolidates the predictions from various base models. Each base model's output is given a corresponding weight, and these weighted predictions are summed along with a bias term to produce a combined decision score. This score is then passed through a non-linear function to yield a probability that indicates the confidence of the final classification.

\begin{equation}
    \tilde{y} = \phi \left( \sum_{r=1}^{R} \theta_r f_r(x) + \delta \right)
\end{equation}

\begin{equation}
    \phi(s) = \frac{1}{1 + e^{-s}}
\end{equation}

\begin{equation}
    \mathcal{L} = -\frac{1}{M} \sum_{m=1}^{M} \left[ t_m \log(\tilde{t}_m) + (1 - t_m) \log(1 - \tilde{t}_m) \right]
\end{equation}

The sigmoid function \cite{Shanthi2022}, denoted by \( \phi(s) \), transforms the aggregate score into a probability ranging from 0 to 1. The learning process involves minimizing the loss function \( \mathcal{L} \), which quantifies the discrepancy between predicted values \( \tilde{t}_m \) and actual labels \( t_m \). This optimization step updates the meta-learner’s weights \( \theta_r \) and bias \( \delta \) to enhance predictive performance.

\subsection{Model Evaluation Metrics}
To assess the prediction performance, this study employed the indices accuracy \cite{powers2020evaluationprecisionrecallfmeasure}, precision, recall, and F1-score. The computation formula for each evaluation index is shown below:

\textbf{ Accuracy:} Accuracy measures the proportion of total correct predictions made by the model.
\begin{equation}
\text{Accuracy} = \frac{TP + TN}{TP + TN + FP + FN}
\end{equation}

\textbf{ Precision:} Precision shows how many of the predicted positive samples were correct.
\begin{equation}
\text{Precision} = \frac{TP}{TP + FP}
\end{equation}

\textbf{ Recall:} Recall measures how many actual positive samples were correctly identified.
\begin{equation}
\text{Recall} = \frac{TP}{TP + FN}
\end{equation}

\textbf{ F1-score:} F1-score is the harmonic mean of precision and recall. It provides a balanced metric, especially useful when classes are imbalanced.
\begin{equation}
\text{F1-score} = 2 \times \frac{\text{Precision} \times \text{Recall}}{\text{Precision} + \text{Recall}}
\end{equation}

Here, $TP$, $TN$, $FP$, and $FN$ represent true positives, true negatives, false positives, and false negatives, respectively.

\section{\label{sec: Result}Results}

\begin{figure}[htbp]
    \centering
    \includegraphics[width=1\columnwidth]{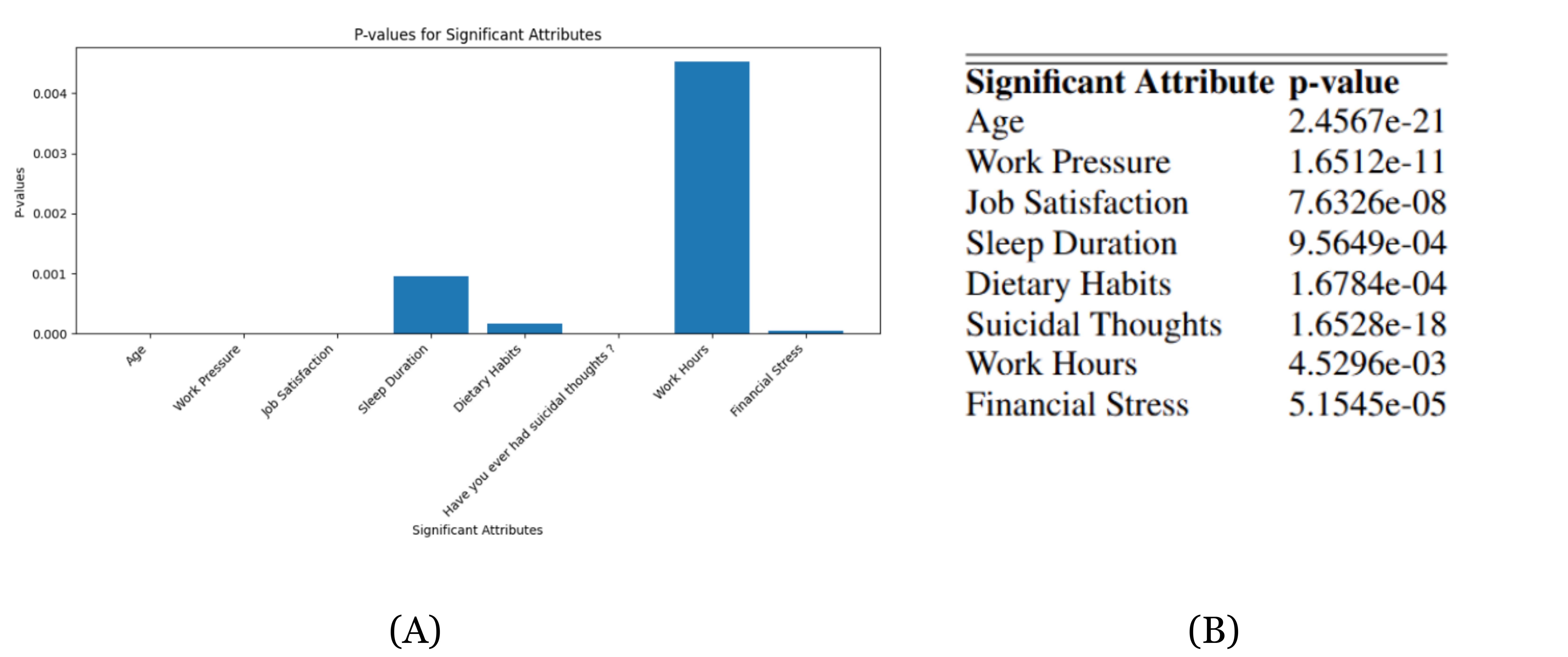}
    \caption{Significant Attributes and their p-value }
    \label{p value}
\end{figure}

The analysis in figure \ref{p value} identified several attributes with statistically highly significant associations, as indicated by their \emph{p}-values which is shown in Figure 2. Age demonstrated the strongest significance ($p = 2.46 \times 10^{-21}$), followed by Suicidal Thoughts ($p = 1.65 \times 10^{-18}$) and Work Pressure ($p = 1.65 \times 10^{-11}$). Other notable attributes included Job Satisfaction ($p = 7.63 \times 10^{-8}$), Dietary Habits ($p = 1.68 \times 10^{-4}$), Financial Stress ($p = 5.15 \times 10^{-5}$), Sleep Duration ($p = 9.56 \times 10^{-4}$), and Work Hours ($p = 4.53 \times 10^{-3}$). Age was further associated with sub-attributes such as Work Pressure, Job Satisfaction, Sleep Duration, Dietary Habits, Work Hours, and Financial Stress, highlighting its central role in the study context.

\noindent

\subsection{AUC-ROC Curve Analysis}
\begin{figure}[htbp]
    \centering
    \includegraphics[width=0.85\linewidth]{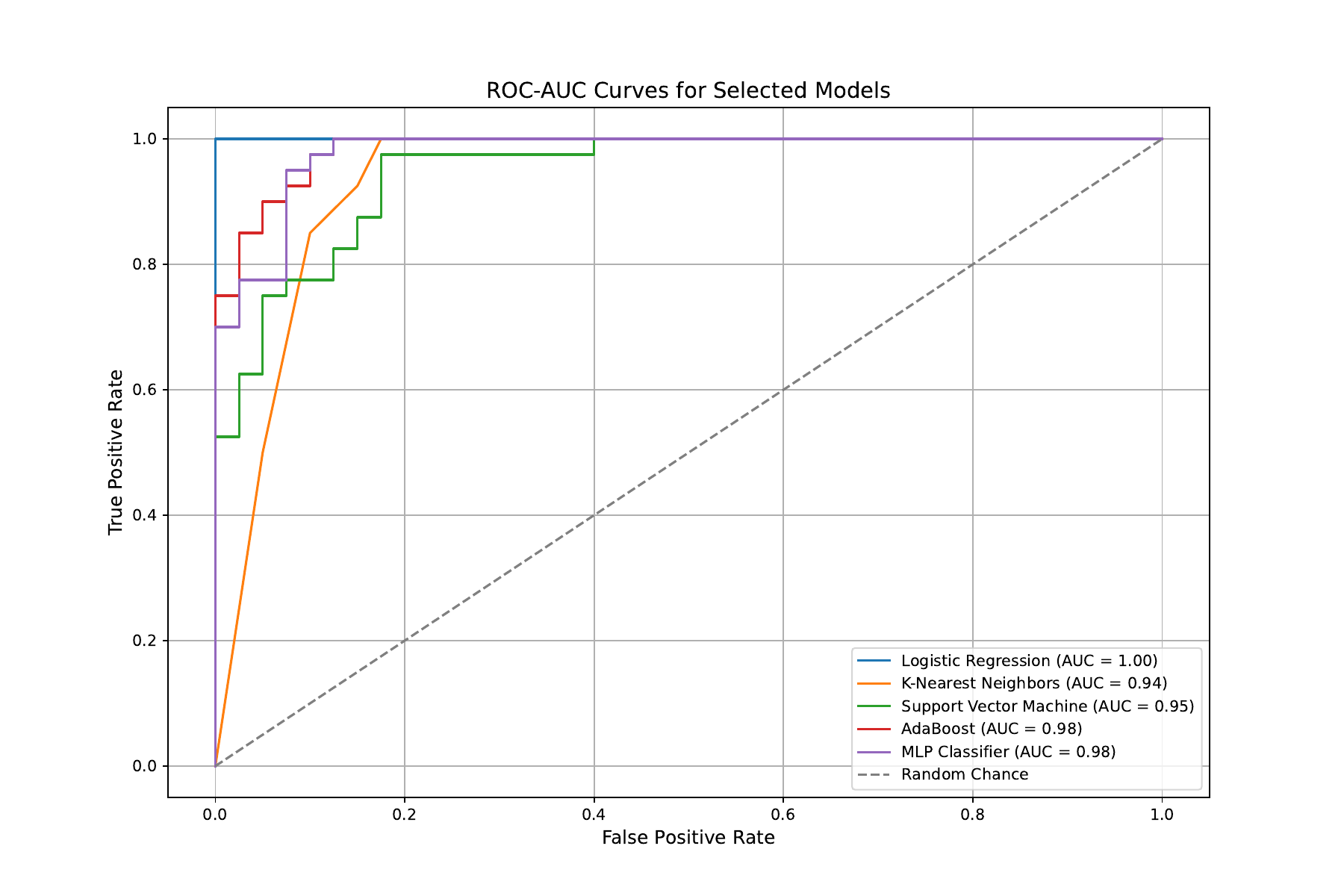}
    \caption{AUC-ROC curve of the selected models}
    \label{fig:roc}
\end{figure}

From Figure \ref{fig:roc}, the AUC-ROC analysis \cite{Jaskowiak2022} on the test dataset demonstrates strong classification performance across the evaluated models. Logistic Regression achieved a perfect AUC score of 1.00, indicating flawless separability between the positive and negative classes. Both AdaBoost and MLP Classifier closely followed with AUC values of 0.98, reflecting highly reliable performance.
Support Vector Machine and K-Nearest Neighbors also showed commendable results, attaining AUC scores of 0.95 and 0.94, respectively. These findings highlight the robustness and generalization ability of the selected models, particularly in binary classification tasks.

\subsection{Confusion Matrix}

\begin{figure}[htbp]
    \centering
    \includegraphics[width=0.7\linewidth]{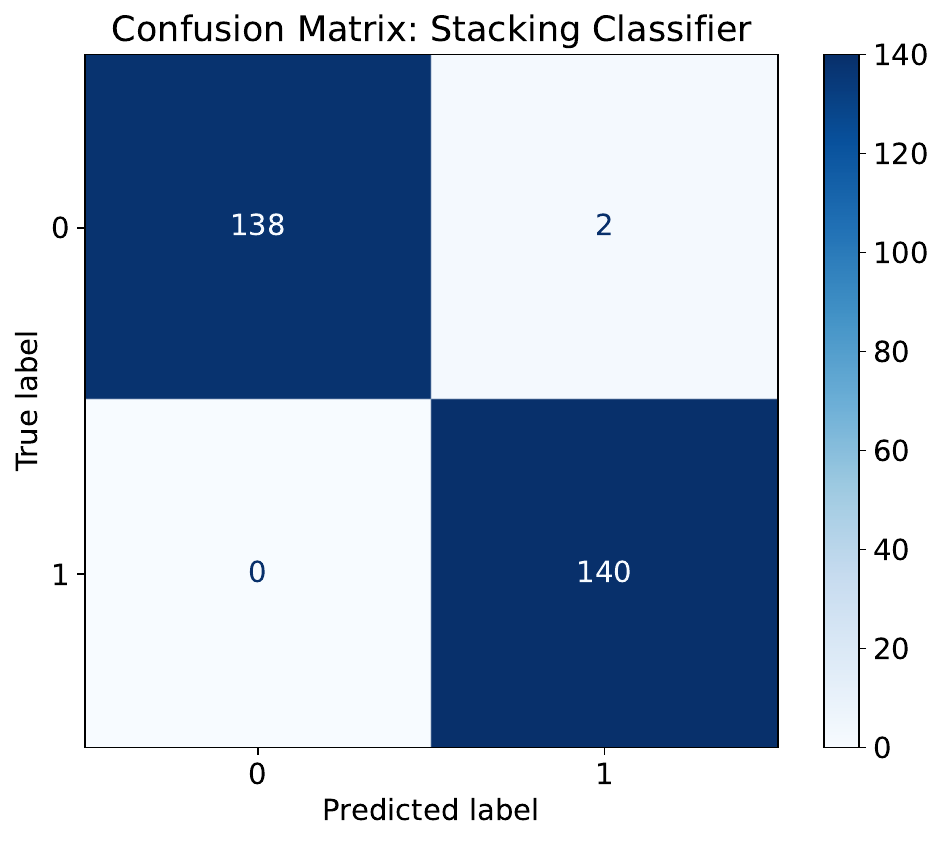}
    \caption{Confusion Matrix }
    \label{fig:conf_matrix}
\end{figure}

Figure \ref{fig:conf_matrix} highlights the effectiveness of the stacking ensemble model, which integrates K-Nearest Neighbors, Support Vector Machine, Multi-Layer Perceptron, and AdaBoost as its base classifiers, with Logistic Regression serving as the meta-classifier. The corresponding confusion matrix \cite{Düntsch_2019} further supports this observation by illustrating the model's strong ability to correctly distinguish between the two classes (Actual 0 and Actual 1). Instances of misclassification were minimal, indicating a well-generalized model that maintains consistency across various input patterns. The detailed numerical results in the matrix underscore the robustness of this ensemble strategy. 

\subsection{Performance Evaluation and Comparison}
The table \ref{tab:model_performance} presents a comprehensive performance comparison of eight machine learning models, evaluated using four widely recognized classification metrics: Accuracy, Precision, Recall, and F1-Score. Among the individual models, Logistic Regression demonstrated remarkably strong and consistent performance, achieving an accuracy of 97.50\%. The Multi-Layer Perceptron (MLP) classifier followed closely, with an accuracy of 93.75\%, showcasing its capability to capture complex, non-linear relationships in the data.

Support Vector Machine (SVM) and AdaBoost produced nearly identical results, each attaining an accuracy of 92.50\%, indicating their robustness and adaptability in handling classification tasks. The K-Nearest Neighbors (KNN) model performed slightly lower, with an accuracy of 91.43\%, likely due to its sensitivity to local data distributions. Gradient Boosting achieved an accuracy of 88.75\%, while the Naïve Bayes classifier yielded the lowest performance, with an accuracy of 86.25\%, suggesting its assumptions were not well-suited to the dataset's characteristics.

Notably, the Stacking Ensemble model significantly outperformed all individual models. By integrating KNN, SVM, MLP, and AdaBoost as base learners and using Logistic Regression as the meta-classifier, the ensemble achieved the highest accuracy of 98.75\%. It also excelled across all other metrics, with a precision of 98.78\%, recall of 98.75\%, and F1-score of 98.75\%.

\begin{table}[htbp]
\caption{Performance Comparison of Classification Models}
\label{tab:model_performance}
\centering
\scriptsize
\setlength{\tabcolsep}{4pt}
\begin{tabularx}{\linewidth}{|X|c|c|c|c|}
\hline
\textbf{Model} & \textbf{Acc.} & \textbf{Prec.} & \textbf{Recall} & \textbf{F1} \\
\hline
Logistic Regression & 97.50 & 97.62 & 97.50 & 97.50 \\
\hline
K-Nearest Neighbors & 91.43 & 92.12 & 91.43 & 91.39 \\
\hline
SVM & 92.50 & 92.61 & 92.50 & 92.51 \\
\hline
Gradient Boosting & 88.75 & 89.37 & 88.75 & 88.71 \\
\hline
AdaBoost & 92.50 & 93.48 & 92.50 & 92.46 \\
\hline
Naive Bayes & 86.25 & 86.46 & 86.25 & 86.23 \\
\hline
MLP Classifier & 93.75 & 94.44 & 93.75 & 93.73 \\
\hline
\textbf{Stacking Ensemble (Prop.)} & \textbf{98.75} & \textbf{98.78} & \textbf{98.75} & \textbf{98.75} \\
\hline
\end{tabularx}
\end{table}

\section{\label{Conclusion}Conclusion}
In this study, we aim at the challenge of predicting depression risk among working professionals by developing a robust stacked ensemble model. Our approach integrates four diverse machine learning algorithms K-Nearest Neighbors, Support Vector Machines, Multi-Layer Perceptron, AdaBoost and combined through logistic regression to leverage their complementary strengths. We ensured that the model focused on the most relevant predictors by using data preprocessing, class balancing, and chi-square-based selection. The final model achieved an impressive 98.75\% accuracy on unseen test data with precision, recall, and F1-scores all exceeding 98\%. Beyond the metrics, our analysis identified critical predictors age, suicidal thoughts, and work pressure as key indicators of depression risk. Future research can explore integrating explainable AI (XAI) techniques, which help interpret the decision-making process of the model, making it more transparent for mental health practitioners. Finally, deploying this system in real-world settings—such as workplace wellness platforms or telehealth applications—could assess its practical impact and usability in supporting early mental health interventions.

\vspace{12pt}

\bibliographystyle{IEEEtran}
\bibliography{Ref}

\begin{thebibliography}{10}
\providecommand{\url}[1]{#1}
\csname url@samestyle\endcsname
\providecommand{\newblock}{\relax}
\providecommand{\bibinfo}[2]{#2}
\providecommand{\BIBentrySTDinterwordspacing}{\spaceskip=0pt\relax}
\providecommand{\BIBentryALTinterwordstretchfactor}{4}
\providecommand{\BIBentryALTinterwordspacing}{\spaceskip=\fontdimen2\font plus
\BIBentryALTinterwordstretchfactor\fontdimen3\font minus \fontdimen4\font\relax}
\providecommand{\BIBforeignlanguage}[2]{{%
\expandafter\ifx\csname l@#1\endcsname\relax
\typeout{** WARNING: IEEEtran.bst: No hyphenation pattern has been}%
\typeout{** loaded for the language `#1'. Using the pattern for}%
\typeout{** the default language instead.}%
\else
\language=\csname l@#1\endcsname
\fi
#2}}
\providecommand{\BIBdecl}{\relax}
\BIBdecl

\bibitem{PRIYA20201258}
A.~Priya, S.~Garg, and N.~P. Tigga, ``Predicting anxiety, depression and stress in modern life using machine learning algorithms,'' \emph{Procedia Computer Science}, vol. 167, pp. 1258--1267, 2020, international Conference on Computational Intelligence and Data Science.

\bibitem{aleem2022depression}
S.~Aleem, N.~Huda, R.~Amin, S.~Khalid, S.~S. Alshamrani, and A.~Alshehri, ``Machine learning algorithms for depression: Diagnosis, insights, and research directions,'' \emph{Electronics}, vol.~11, no.~7, p. 1111, 2022.

\bibitem{ZULFIKER2021100044}
M.~S. Zulfiker, N.~Kabir, A.~A. Biswas, T.~Nazneen, and M.~S. Uddin, ``An in-depth analysis of machine learning approaches to predict depression,'' \emph{Current Research in Behavioral Sciences}, vol.~2, p. 100044, 2021.

\bibitem{PATEL2016115}
M.~J. Patel, A.~Khalaf, and H.~J. Aizenstein, ``Studying depression using imaging and machine learning methods,'' \emph{NeuroImage: Clinical}, vol.~10, pp. 115--123, 2016.

\bibitem{chekroud2016cross}
A.~M. Chekroud, R.~J. Zotti, Z.~Shehzad, R.~Gueorguieva, M.~K. Johnson, M.~H. Trivedi, T.~D. Cannon, J.~H. Krystal, and P.~R. Corlett, ``Cross-trial prediction of treatment outcome in depression: A machine learning approach,'' \emph{The Lancet Psychiatry}, vol.~3, no.~3, pp. 243--250, 2016.

\bibitem{LEE2018519}
Y.~Lee, R.-M. Ragguett, R.~B. Mansur, J.~J. Boutilier, J.~D. Rosenblat, A.~Trevizol, E.~Brietzke, K.~Lin, Z.~Pan, M.~Subramaniapillai, T.~C. Chan, D.~Fus, C.~Park, N.~Musial, H.~Zuckerman, V.~C. Chen, R.~Ho, C.~Rong, and R.~S. McIntyre, ``Applications of machine learning algorithms to predict therapeutic outcomes in depression: A meta-analysis and systematic review,'' \emph{Journal of Affective Disorders}, vol. 241, pp. 519--532, 2018.

\bibitem{Sajjadian2021}
M.~Sajjadian, R.~W. Lam, R.~Milev, S.~Rotzinger, B.~N. Frey, C.~N. Soares, S.~V. Parikh, J.~A. Foster, G.~Turecki, D.~J. Müller, and et~al., ``Machine learning in the prediction of depression treatment outcomes: a systematic review and meta-analysis,'' \emph{Psychological Medicine}, vol.~51, no.~16, pp. 2742--2751, 2021.

\bibitem{LI2019101696}
X.~Li, X.~Zhang, J.~Zhu, W.~Mao, S.~Sun, Z.~Wang, C.~Xia, and B.~Hu, ``Depression recognition using machine learning methods with different feature generation strategies,'' \emph{Artificial Intelligence in Medicine}, vol.~99, p. 101696, 2019.

\bibitem{Lee2022}
K.-S. Lee and B.-J. Ham, ``Machine learning on early diagnosis of depression,'' \emph{Psychiatry Investigation}, vol.~19, no.~8, pp. 597--605, 2022.

\bibitem{Richter2020}
T.~Richter, B.~Fishbain, A.~Markus, G.~Richter-Levin, and H.~Okon-Singer, ``Using machine learning-based analysis for behavioral differentiation between anxiety and depression,'' \emph{Scientific Reports}, vol.~10, no.~1, p. 16381, 2020.

\bibitem{KUMAR20201989}
P.~Kumar, S.~Garg, and A.~Garg, ``Assessment of anxiety, depression and stress using machine learning models,'' \emph{Procedia Computer Science}, vol. 171, pp. 1989--1998, 2020, third International Conference on Computing and Network Communications (CoCoNet'19).

\bibitem{Jain_2022}
P.~Jain, K.~Ram~Srinivas, and A.~Vichare, ``Depression and suicide analysis using machine learning and nlp,'' \emph{Journal of Physics: Conference Series}, vol. 2161, no.~1, p. 012034, jan 2022.

\bibitem{sharma2023depression}
S.~Sharma, ``Depression survey/dataset for analysis,'' \url{https://www.kaggle.com/datasets/sumansharmadataworld/depression-surveydataset-for-analysis}, 2023, accessed: May 31, 2025.

\bibitem{SosaCabrera2024}
G.~Sosa-Cabrera, S.~Gómez-Guerrero, M.~García-Torres, and C.~E. Schaerer, ``Feature selection: a perspective on inter-attribute cooperation,'' \emph{International Journal of Data Science and Analytics}, vol.~17, no.~2, pp. 139--151, Mar 2024.

\bibitem{Byeon2017CHAID}
H.~Byeon, ``Chi-square automatic interaction detection modeling for predicting depression in multicultural female students,'' \emph{International Journal of Advanced Computer Science and Applications (IJACSA)}, vol.~8, no.~12, pp. 179--183, 2017.

\bibitem{PONOMARENKO2013290}
``Degrees of freedom,'' in \emph{Brenner's Encyclopedia of Genetics (Second Edition)}, second edition~ed., S.~Maloy and K.~Hughes, Eds.\hskip 1em plus 0.5em minus 0.4em\relax San Diego: Academic Press, 2013, pp. 290--292.

\bibitem{zhang2024functionallinearregressioncumulative}
Q.~Zhang, A.~Makur, and K.~Azizzadenesheli, ``Functional linear regression of cumulative distribution functions,'' 2024.

\bibitem{Guo2003}
G.~Guo, H.~Wang, D.~Bell, Y.~Bi, and K.~Greer, ``Knn model-based approach in classification,'' in \emph{On The Move to Meaningful Internet Systems 2003: CoopIS, DOA, and ODBASE}, R.~Meersman, Z.~Tari, and D.~C. Schmidt, Eds.\hskip 1em plus 0.5em minus 0.4em\relax Berlin, Heidelberg: Springer Berlin Heidelberg, 2003, pp. 986--996.

\bibitem{Suthaharan2016}
S.~Suthaharan, \emph{Support Vector Machine}.\hskip 1em plus 0.5em minus 0.4em\relax Boston, MA: Springer US, 2016, pp. 207--235.

\bibitem{MURTAGH1991183}
F.~Murtagh, ``Multilayer perceptrons for classification and regression,'' \emph{Neurocomputing}, vol.~2, no.~5, pp. 183--197, 1991.

\bibitem{Schapire2013}
R.~E. Schapire, ``Explaining adaboost,'' in \emph{Empirical Inference: Festschrift in Honor of Vladimir N. Vapnik}, B.~Sch{\"o}lkopf, Z.~Luo, and V.~Vovk, Eds.\hskip 1em plus 0.5em minus 0.4em\relax Berlin, Heidelberg: Springer Berlin Heidelberg, 2013, pp. 37--52.

\bibitem{sperandei2014understanding}
S.~Sperandei, ``Understanding logistic regression analysis,'' \emph{Biochemia Medica (Zagreb)}, vol.~24, no.~1, pp. 12--18, 2014.

\bibitem{Shanthi2022}
D.~L. Shanthi and N.~Chethan, ``Genetic algorithm based hyper-parameter tuning to improve the performance of machine learning models,'' \emph{SN Computer Science}, vol.~4, no.~2, p. 119, Dec 2022.

\bibitem{powers2020evaluationprecisionrecallfmeasure}
D.~M.~W. Powers, ``Evaluation: from precision, recall and f-measure to roc, informedness, markedness and correlation,'' 2020.

\bibitem{Jaskowiak2022}
P.~A. Jaskowiak, I.~G. Costa, and R.~J. G.~B. Campello, ``The area under the roc curve as a measure of clustering quality,'' \emph{Data Mining and Knowledge Discovery}, vol.~36, no.~3, pp. 1219--1245, 2022.

\bibitem{Düntsch_2019}
I.~Düntsch and G.~Gediga, ``Confusion matrices and rough set data analysis,'' \emph{Journal of Physics: Conference Series}, vol. 1229, no.~1, p. 012055, may 2019.

\end{thebibliography}

\end{document}